\documentclass[10pt,twocolumn,letterpaper]{article}

\usepackage{cvpr}              

\usepackage[accsupp]{axessibility}
\usepackage{graphicx}
\usepackage{amsmath}
\usepackage{amssymb}
\usepackage{booktabs}
\usepackage{pifont}
\usepackage{bm}

\makeatletter
\renewcommand{\maketag@@@}[1]{\hbox{\m@th\normalsize\normalfont#1}}%
\makeatother

\usepackage[pagebackref,breaklinks,colorlinks]{hyperref}

\usepackage[capitalize]{cleveref}
\crefname{section}{Sec.}{Secs.}
\Crefname{section}{Section}{Sections}
\Crefname{table}{Table}{Tables}
\crefname{table}{Tab.}{Tabs.}

\def\cvprPaperID{2775} 
\def\confName{CVPR}
\def\confYear{2023}

\begin{document}

\title{IPCC-TP: Utilizing Incremental Pearson Correlation Coefficient\\ for Joint Multi-Agent Trajectory Prediction}

\author{
\hspace{-12pt}
Dekai Zhu$^{1,\ast}$, 
Guangyao Zhai$^{1,\ast}$, 
Yan Di$^{1,\dag}$, 
Fabian Manhardt$^{2}$,\\
Hendrik Berkemeyer$^{3,4}$,
Tuan Tran$^{3}$,
Nassir Navab$^{1}$,
Federico Tombari$^{1,2}$, and
Benjamin Busam$^{1,5}$\\
\\
$^1$ Technical University of Munich\quad
$^2$ Google \\
$^3$ Robert Bosch GmbH \quad
$^4$ University of Osnabrueck \quad
$^5$ 3Dwe.ai \\
\small{\texttt{\{firstname.lastname,b.busam\}@tum.de}} \\
}
\maketitle

\begin{abstract}

Reliable multi-agent trajectory prediction is crucial for the safe planning and control of autonomous systems.
Compared with single-agent cases, the major challenge in simultaneously processing multiple agents lies in modeling complex social interactions caused by various driving intentions and road conditions.
Previous methods typically leverage graph-based message propagation or attention mechanism to encapsulate such interactions in the format of marginal probabilistic distributions.
However, it is inherently sub-optimal.
In this paper, we propose IPCC-TP, a novel relevance-aware module based on \textbf{I}ncremental \textbf{P}earson \textbf{C}orrelation \textbf{C}oefficient to improve multi-agent interaction modeling.
IPCC-TP learns pairwise joint Gaussian Distributions through the tightly-coupled estimation of the means and covariances according to interactive incremental movements.
Our module can be conveniently embedded into existing multi-agent prediction methods to extend original motion distribution decoders. 
Extensive experiments on nuScenes and Argoverse 2 datasets demonstrate that IPCC-TP improves the performance of baselines by a large margin.

\end{abstract}

\vspace{-2mm}
{\let\thefootnote\relax\footnote{{$^\ast$ Equal contribution. $^\dag$ Corresponding author.\vspace{-2mm}}}} 
\vspace{-2mm}


\label{sec:intro}
\begin{figure}[t]
    \centering
    \includegraphics[width=0.98\linewidth]{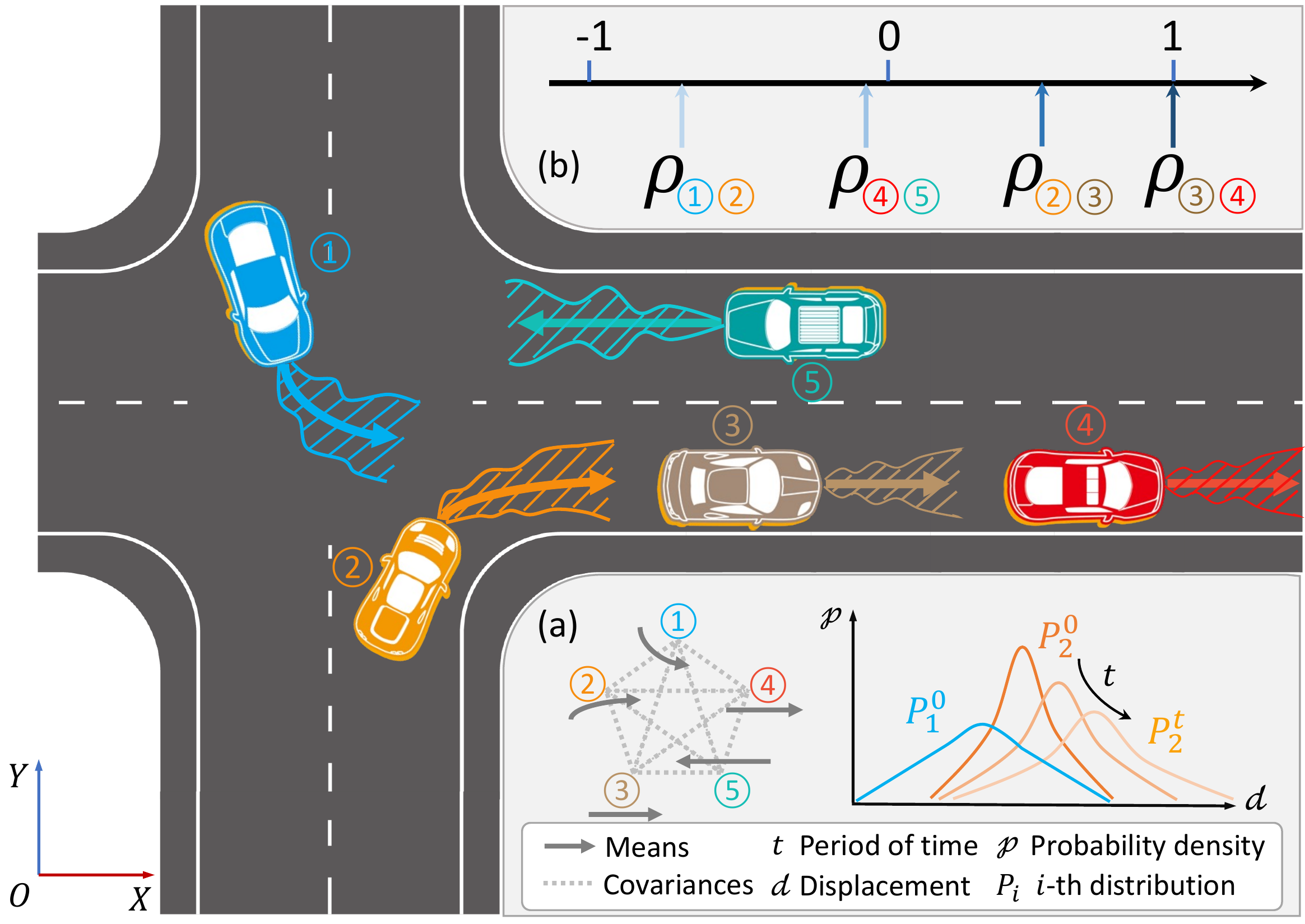}
    \caption{Schematic illustration of a driving scene with five agents \ding{172}$-$\ding{176}. \textbf{(a)} IPCC-TP models future interactions for $t$ steps based on joint Gaussian distribution among one-dimensional increments of agents' displacement $d$.
    \textbf{(b)} The values of IPCC represented by $\rho$ on the axis indicate the correlation between agents' trajectories. In the predicted future, \ding{172} and \ding{173} interact intensively with \ding{172} yielding to \ding{173} ($\rho_{12}<0$). \ding{175}'s movement does not interfere \ding{176} ($|\rho_{45}|\approx 0$). \ding{174} is closely following \ding{175} ($\rho_{34}\leq +1$).} 
    \vspace{-3mm}
    \label{fig:teasor}
\end{figure}


\section{Introduction}
Trajectory prediction refers to predicting the future trajectories of one or several target agents based on past trajectories and road conditions.
It is an essential and emerging subtask in autonomous driving~\cite{su2023opa,liu2019unsupervised,zhai2020flowmot,kong2020semantic} and industrial robotics~\cite{jetchev2009trajectory,rosmann2017online,zhang2021efficient}.

Previous methods~\cite{chai2020multipath,phan2020covernet,zhao2021tnt,gu2021densetnt,fang2020tpnet} concentrate on Single-agent Trajectory Prediction (STP), which leverages past trajectories of other agents and surrounding road conditions as additional cues to assist ego-motion estimation. Despite the significant progress made in recent years, its application is limited. The reason is that simultaneously processing all traffic participants remains unreachable, although it is a common requirement for safe autonomous driving.

A straightforward idea to deal with this issue is to directly utilize STP methods on each agent respectively and finally attain a joint collision-free prediction via pairwise collision detection. 
Such methods are far from reliably handling Multi-agent Trajectory Prediction (MTP) since the search space for collision-free trajectories grows exponentially as the number of agents increases, \textit{i.e.}, $N$ agents with $M$ modes yield $M^N$ possible combinations, making such search strategy infeasible with numerous agents. 
By incorporating graph-based message propagation or attention mechanism into the framework to model the future interactions among multiple agents, \cite{ngiam2021scene,girgis2021latent,yuan2021agentformer} aggregate all available cues to simultaneously predict the trajectories for multiple agents in the format of a set of marginal probability distributions. 
While these frameworks improve the original STP baseline to some extent, the estimation based on a set of marginal distributions is inherently sub-optimal~\cite{ProbabilityPI}.
To address this issue, a predictor that can predict a joint probability distribution for all agents in the scene is necessary.
An intuitive strategy for designing such a predictor is to directly predict all parameters for means and covariance matrices, which define the joint probability distributions~\cite{tang2021collaborative}.
To predict a trajectory of $T$ time-steps with $N$ agents and $M$ modes up to $4N^2TM$ covariance parameters are needed.
However, such redundant parameterization fails to analyze the physical meaning of covariance parameters and also makes it hard to control the invertibility of the covariance matrix.

In this paper, we take a further step to explicitly interpret the physical meanings of the individual variances and Pearson Correlation Coefficients (PCC) inside the covariance matrix.
Instead of directly modeling the locations of agents at each sampled time, we design a novel movement-based method which we term Incremental PCC (IPCC). It helps to simulate the interactions among agents during driving.
IPCC models the increment from the current time to a future state at a specific step.
Thereby individual variances capture the local uncertainty of each agent in the Bird-Eye-View without considering social interactions (Figure~\ref{fig:teasor}.a), while IPCC indicates the implicit driving policies of two agents, e.g., 
one agent follows or yields to another agent or the two agents' motions are irrelevant (Figure~\ref{fig:teasor}.b).
Compared to position-based modeling, IPCC has two advantages.
First, IPCC models social interactions during driving in a compact manner.
Second, since position-based modeling requires respective processes along the two axes, IPCC can directly model the motion vector, which reduces memory cost for covariance matrix storage by a factor of four and facilitates deployment on mobile platforms, such as autonomous vehicles and service robots.

Based on IPCC, we design and implement IPCC-TP, a module for the MTP task. IPCC-TP can be conveniently embedded into existing methods by extending the original motion decoders to boost their performance.
Experiments show that IPCC-TP effectively captures the pairwise relevance of agents and predicts scene-compliant trajectories.

In summary, the main contributions of this work are threefold:
\begin{itemize}
    \item We introduce a novel movement-based method, namely IPCC, that can intuitively reveal the physical meaning of pairwise motion relevance and facilitate deployment by reducing memory cost.
    \item Based on IPCC, we propose IPCC-TP, which is compatible with state-of-the-art MTP methods and possess the ability to model future interactions with joint Gaussian distributions among multiple agents.
    \item Experiment results show that methods enhanced with IPCC-TP outperform the original methods by a large margin and become the new state-of-the-art methods.
    

\end{itemize}

\section{Related Work}

In this section, we first briefly review the general frameworks of learning-based trajectory prediction and then introduce the literature closely related to our approach.


\noindent
\textbf{General trajectory prediction architecture}.
Trajectory prediction for moving agents is important for the safe planning of autonomous driving~\cite{cui2019multimodal, gao2020vectornet, yuan2021agentformer}. One branch of trajectory prediction consists of the approaches~\cite{alahi2016social,mercat2020multi,salzmann2020trajectron,liang2020learning,casas2020spagnn,chai2020multipath,phan2020covernet,zhao2021tnt,gu2021densetnt,fang2020tpnet} based on typical neural networks (CNN, GNN~\cite{kipf2017semi,li2016gated, di2022gpv}, RNN~\cite{hochreiter1997long,Junyoung2014}). 
Cui et al.~\cite{cui2019multimodal} propose to use agents' states and features extracted by CNN from Bird-Eye-View images to generate a multimodal prediction. Similarly, CoverNet~\cite{phan2020covernet} uses CNN to encode the scene context while it proposes dozens of trajectory anchors based on the bicycle model and converts prediction from position regression to anchor classification. 
A hierarchical GNN named VectorNet~\cite{gao2020vectornet} encapsulates the sequential features of map elements and past trajectories with instance-wise subgraphs and models interactions with a global graph.
TNT~\cite{zhao2021tnt} and DenseTNT~\cite{gu2021densetnt} are extensions of~\cite{gao2020vectornet} that focus on predicting reasonable destinations. Social LSTM \cite{alahi2016social} models the trajectories of individual agents from separate LSTM networks and aggregates the LSTM hidden cues to model their interactions.
CL-SGR~\cite{CL_SGR} considers the sample replay model in a continuous trajectory prediction scenario setting to avoid catastrophic forgetting.
The other branch~\cite{girgis2021latent,ngiam2021scene,yuan2021agentformer} models the interaction among the agents based on the attention mechanism. They work with the help of Transformer \cite{vaswani2017attention}, which achieves huge success in the fields of natural language processing~\cite{kenton2019bert} and computer vision~\cite{carion2020end,liu2021swin,dosovitskiy2020image, zhang2022sst,zhai2022monograspnet}. 
Scene Transformer \cite{ngiam2021scene} mainly consists of attention layers,
including self-attention layers that encode sequential features on the temporal dimension, self-attention layers that capture interactions on the social dimension between traffic participants, and cross-attention layers that learn compliance with traffic rules.

\noindent
\textbf{Prediction of multi-agent interaction}.
Most aforementioned methods are proposed for the STP task.
Recently, more work has been concentrated on the MTP task.
M2I~\cite{sun2022m2i} uses heuristic-based labeled data to train a module for classifying the interacting agents as pairs of influencers and reactors. M2I first predicts a marginal trajectory for the influencer and then a conditional trajectory for the reactor. 
But M2I can only deal with interactive scenes consisting of two interacting agents. 
Models such as~\cite{girgis2021latent,yuan2021agentformer,ngiam2021scene} are proposed for general multi-agent prediction.
Scene Transformer~\cite{ngiam2021scene} and AutoBots~\cite{girgis2021latent} share a similar architecture, while the latter employs learnable seed parameters in the decoder to accelerate inference.
AgentFormer~\cite{yuan2021agentformer} leverages the sequential representation of multi-agent trajectories by flattening the state features across agents and time. 
Although these methods utilize attention layers intensively in their decoders to model future interactions and achieve state-of-the-art performance, their predictions are suboptimal since the final prediction is not a joint probability distribution between agents but only consists of marginal probability distributions. 
To the best of our knowledge,~\cite{tang2021collaborative} is the only work that describes the future movement of all agents with a joint probability distribution.
This work investigates the collaborative uncertainty of future interactions, mathematically speaking, the covariance matrix per time step. They use MLP decoders to predict mean values and the inverse of covariance matrices directly. 
We refer to~\cite{tang2021collaborative} as CUM (\textbf{C}ollaborative \textbf{U}ncertainty based \textbf{M}odule) for convenience and provide methodological and quantitative comparisons between CUM and our module in Sec.~\ref{sec:method} and~\ref{experiments}.

\section{Methodology}
\label{sec:method}
In this section, we first introduce the background knowledge of MTP. 
Then we propose the IPCC method, which models the relevance of the increments of target agents.
Based on this, we design IPCC-TP and illustrate how this module can be plugged into the state-of-the-art models~\cite{yuan2021agentformer,girgis2021latent}.
In the end, we compare our module with CUM.

\subsection{Preliminary}
\label{preliminary}
\noindent
\textbf{Problem Statement.}
Given the road map $\mathcal{G}$ and $N$ agents with trajectories of past $T_{obs}$ steps, MTP is formulated as predicting future trajectories of all agents for $T$ steps.
For observed time steps $t \in (-T_{obs}, 0]$, the past states at $t$ step are $\mathcal{H}_t = \{ \mathcal{H}^t_i | i=1,...,N \}$, where $\mathcal{H}^t_i$ contains the state information, \textit{i.e.} $\{x, y\}$ position of agent $i$ at this time step. 
And the past trajectories are denoted as $\mathcal{H} = \{ \mathcal{H}_t | -T_{obs} < t \leq 0 \}$. 
Similarly, for future time steps $t \in (0, T]$, the future states at step $t$ are $\mathcal{F}_t = \{ \mathcal{F}^t_i | i=1,...,N\}$, where $\mathcal{F}^t_i$ contains $\{x, y\}$ position of agent $i$ at this step.
And the future trajectories are $\mathcal{F} = \{ \mathcal{F}_t | 0 < t \leq T \}$. 
Considering that a specific current state could develop to multiple interaction patterns, most MTP methods predict $M$ interaction modes in the future. 
For simplicity, we only analyze the cases of $M=1$ in this section and our method can be conveniently extended to predict multiple modes. 

\smallskip
\noindent
\textbf{Marginal Probability Modeling.}
In learning-based MTP methods, the training objective is to find the optimal parameter $\theta^*$ to maximize the Gaussian likelihood $P$:
\begin{equation}
\theta^*=\arg \max_{\theta} P\left(\mathcal{F} \mid \mathcal{H}, \mathcal{G}, \theta\right).
\label{eq:obj}
\end{equation}

Thereby the core of MTP lies in constructing an effective joint probability distribution $P$ to model the generative future trajectories of all agents.
For simplicity, existing methods such as AutoBots \cite{girgis2021latent} and AgentFormer \cite{yuan2021agentformer} approximate the high-dimensional multivariate distribution $P=\{P_i^t\}$ in Eq.~\eqref{eq:obj} by aggregating multiple marginal probability distributions.
Each distribution $P_i^t$ describes $i$-th agent's position at a specific time $t$. 
The objective function Eq.~\eqref{eq:obj} is then transformed as:
\begin{equation}
\theta^*=\arg \max_{\theta} \sum_{t=1}^{T} \sum_{i=1}^N P_i^t\left(\mathcal{F}_i^t \mid \mathcal{H}, \mathcal{G}, \theta\right),
\end{equation}
where the marginal prediction $P_i^t \sim\mathcal{N}(\mu_i^t, \Sigma_i^t)$ is a 2D Gaussian distribution defined by:
\begin{equation}
	\begin{aligned}
	\mu_i^t &= [\mu^{tx}_i, \: \mu^{ty}_i] \\[1em]
	\Sigma_i^t &= \begin{bmatrix}
		(\sigma^{tx}_i)^2 & \rho^{txy}_{i} \sigma^{tx}_i \sigma^{ty}_i\\[1em]
		\rho^{tyx}_{i} \sigma^{ty}_i \sigma^{tx}_i & (\sigma^{ty}_i)^2
	\end{bmatrix},
\end{aligned}
\end{equation}
where $\mu_i^t$ denotes the mean position of the $i$-th agent at time step $t$, $\Sigma_i^t$ is its corresponding covariance matrix.
$\rho^{txy}_{i}, \rho^{tyx}_{i}$ in $\Sigma_i^t$ are the PCCs of agent $i$ on $x,y$ axis respectively.

However, such marginal approximation is inherently sub-optimal~\cite{ProbabilityPI}, resulting in unsatisfactory results on real-world benchmarks~\cite{caesar2020nuscenes,chang2019argoverse,pellegrini2009you,lerner2007crowds,kothari2021human}.
Although existing methods typically employ attention-based mechanism to help each agent attends to its interactive counterparts, social interactions in motion between different pairs of agents is in fact not explicitly analyzed and modeled, which limits the capability of networks to predict scene compliant and collision-free trajectories.

\subsection{PCC: Position Correlation Modeling}
\label{general}
The joint probability distribution among multiple agents in the scene contains correlations between the motions of different agents, which are essential parts of precisely modeling future interactions. 
Theoretically, using such a scene-level modeling method would lead to a more optimal estimation of future interactions\cite{ProbabilityPI} than stacking a set of marginal probability distributions to approximate the joint distribution.  

For time step  $t$, the joint prediction among agents in the scene $\hat{\mathcal{F}_t}=P_t\left(\cdot \mid \mathcal{H}, \mathcal{M}, \theta\right)\sim\mathcal{N}(M_{t}, \Sigma_{t})$ is defined by:
\begin{equation}
M_{t}=\left[\mu_1^t, \ldots, \mu_N^t\right],\\ \mu_i^t=\left[\mu_i^{tx}, \mu_i^{ty}\right],
\label{eq:joint_mean}
\end{equation}
\begin{equation}
\begin{gathered}
\Sigma_{t}=
\left[\begin{array}{ccc}
{\sigma_{11}^t} & \cdots & {\sigma_{1 N}^t} \\
\vdots & \ddots & \vdots \\
{\sigma_{N 1}^t} & \cdots & {\sigma_{N N}^t}
\end{array}\right],\\[1em]
{\sigma_{ij}^t}=\left[\begin{array}{cc}
\rho_{ij}^{txx}\sigma_{i}^{tx}\sigma_{j}^{tx} & \rho_{ij}^{txy}\sigma_{i}^{tx}\sigma_{j}^{ty} \\[1em]
\rho_{ij}^{tyx}\sigma_{i}^{ty}\sigma_{j}^{tx} & \rho_{ij}^{tyy}\sigma_{i}^{ty}\sigma_{j}^{ty}
\end{array}\right],
\end{gathered}
\label{eq:joint_cov}
\end{equation}
where $M_{t} \in \mathbb{R}^{2 N}$ contains the mean positions of all agents, and $\Sigma_{t} \in \mathbb{R}^{2 N \times 2 N}$ is the corresponding covariance matrix, and $\rho_{ij}^{txx}$ = $\rho_{ij}^{tyy} = 1$ when $i=j$. 
The objective function Eq.~\eqref{eq:obj} then turns into:
\begin{equation}
\theta^*=\arg \max_{\theta} \sum_{t=1}^{T} P_t\left(\mathcal{F}_t \mid \mathcal{H}, \mathcal{G}, \theta\right).
\end{equation}

Thereby, a straightforward strategy for designing a predictor based on such joint probability modeling is to directly predict all parameters in means and covariance matrices.
However, such redundant parameterization still fails to capture the interaction patterns in motion between agents since PCCs inside the covariance matrix reflect the \textit{agent-to-agent} position correlations rather than motion correlations.
Moreover, such strategy also makes it hard to control the invertibility of the covariance matrix.

\begin{figure}[t]
    \centering
    \includegraphics[width=1.00\linewidth]{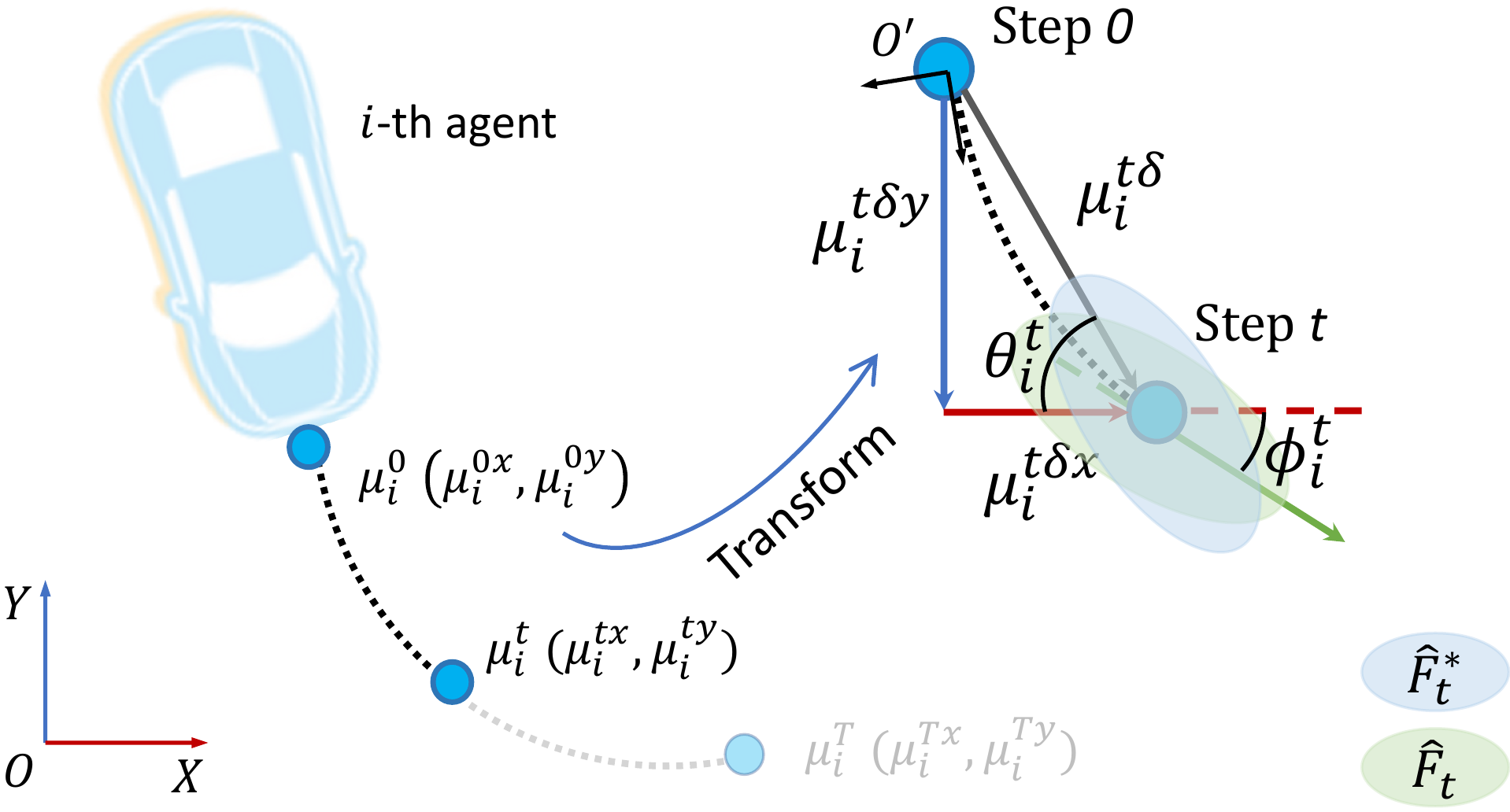}
    \caption{The movement from position $\mu^0_i$ to $\mu^t_i$ in the x-y coordinate system corresponds to an incremental movement of $\mu^{t \delta}_i$ in IPCC method (solid black line: incremental movement, block dash line: actual trajectory). $\theta^t_i$ is the approximate yaw angle, while $\phi^t_i$ is the actual yaw angle. When $\theta^t_i$ is close to $\phi^t_i$, the prediction $\hat{\mathcal{F}^{*}_t}$ derived from IPCC method is an approximation to $\hat{\mathcal{F}_t}$ in the x-y coordinate system.} 
    \label{fig:increment}
\end{figure}

\subsection{IPCC: Motion Interaction Modeling.}
\label{IPCC}

Since the joint Gaussian distribution defined by Eq.~\eqref{eq:joint_mean} and Eq.~\eqref{eq:joint_cov} is not feasible in real applications, we propose a novel motion interaction modeling method, IPCC.
For Eq.~\eqref{eq:joint_cov}, we observe that each pair of agents requires estimations of PCCs: $\{\rho_{ij}^{txx}, \rho_{ij}^{txy}, \rho_{ij}^{tyx}, \rho_{ij}^{tyy}\}$. 
These representations are redundant since they respectively model the pairwise position correlation between agent $i$ and $j$ along the 2D axis from the bird-eye view.
In the following part, we demonstrate that this correlation can be merged into a single motion-aware parameter IPCC by modeling the one-dimensional increments from current time step (t=0) to any future time step.

\smallskip
\noindent
\textbf{From PCC to IPCC.} 
For future time step $t > 0$, we predict the increments $\hat{\mathcal{F}_{\Delta}^t}$ as shown in Fig.~\ref{fig:increment}, which is the one-dimensional displacements with respect to the current positions $M_{0}$, for all agents in the scene. 
$\hat{\mathcal{F}_{\Delta}^t} \sim \mathcal{N}(M_{\Delta}^t, \Sigma_{\Delta}^t)$ is defined by:
\begin{equation}
M_{\Delta}^t =\left[\mu^{t \delta}_1, \ldots, \mu^{t \delta}_N\right], \mu^{t \delta}_i \in \mathbb{R}^{+}_{0} 
\end{equation}
\begin{equation}
\begin{aligned}
\Sigma_{\Delta}^t = P_{\Delta}^t \odot &\left[\begin{array}{ccc}
\left(\sigma^{t \delta}_1\right)^2 & \ldots & \sigma^{t \delta}_1 \sigma^{t \delta}_N \\
\vdots & \ddots & \vdots \\
\sigma^{t \delta}_N \sigma_1^{t \delta} & \cdots & \left(\sigma_N^{t \delta}\right)^2
\end{array}\right],\\[1em]
P_\Delta^t = &\left[\begin{array}{ccc}
1 & \ldots & \rho^{t \delta}_{1 N}\\
\vdots & \ddots & \vdots \\
\rho^{t \delta}_{N 1} & \ldots & 1
\end{array}\right],
\end{aligned}
\label{eq:Ptd}
\end{equation}
where $P_\Delta^t, \Sigma_{\Delta}^t \in \mathbb{R}^{N \times N}$. 
$\Sigma_{\Delta}^t$ denotes the covariance matrix and IPCC matrix $P_\Delta^t$ can be extracted from $\Sigma_{\Delta}^t$ as in Eq.~\eqref{eq:Ptd}.
$P_\Delta^t$ preserves all pairwise correlation parameters for the increments.
$\odot$ denotes element-wise multiplication.
Thereby $\rho^{t \delta}_{i j}\in [-1, +1]$ naturally encapsulates the motion relevance between $i$-th and $j$-th agents, which conveys various interaction patterns: 
\textbf{(1)} The more it approaches $+1$, the more similar behaviours the two agents share, \textit{e.g.}, two vehicles drive on the same lane towards the same direction. 
\textbf{(2)} The more it approaches $-1$, the more one vehicle is likely to yield if the other one is driving aggressively, \textit{e.g.}, two vehicles are at a merge.
\textbf{(3)} The closer it is to $0$ the less two agents impact each other.
Such explicit modeling of motion interactions enforces the network to effectively predict the trajectory of each agent by attending to different interaction patterns from other agents.

Since we aim to build a plug-in module that can be conveniently embedded into other existing methods, we project the predicted IPCC parameters back to the original $x,y$ axis.
Therefore, the training losses of baselines can be directly inherited and extended for supervision.

\begin{figure*}[t]
    \centering
    \includegraphics[width=0.98\linewidth]{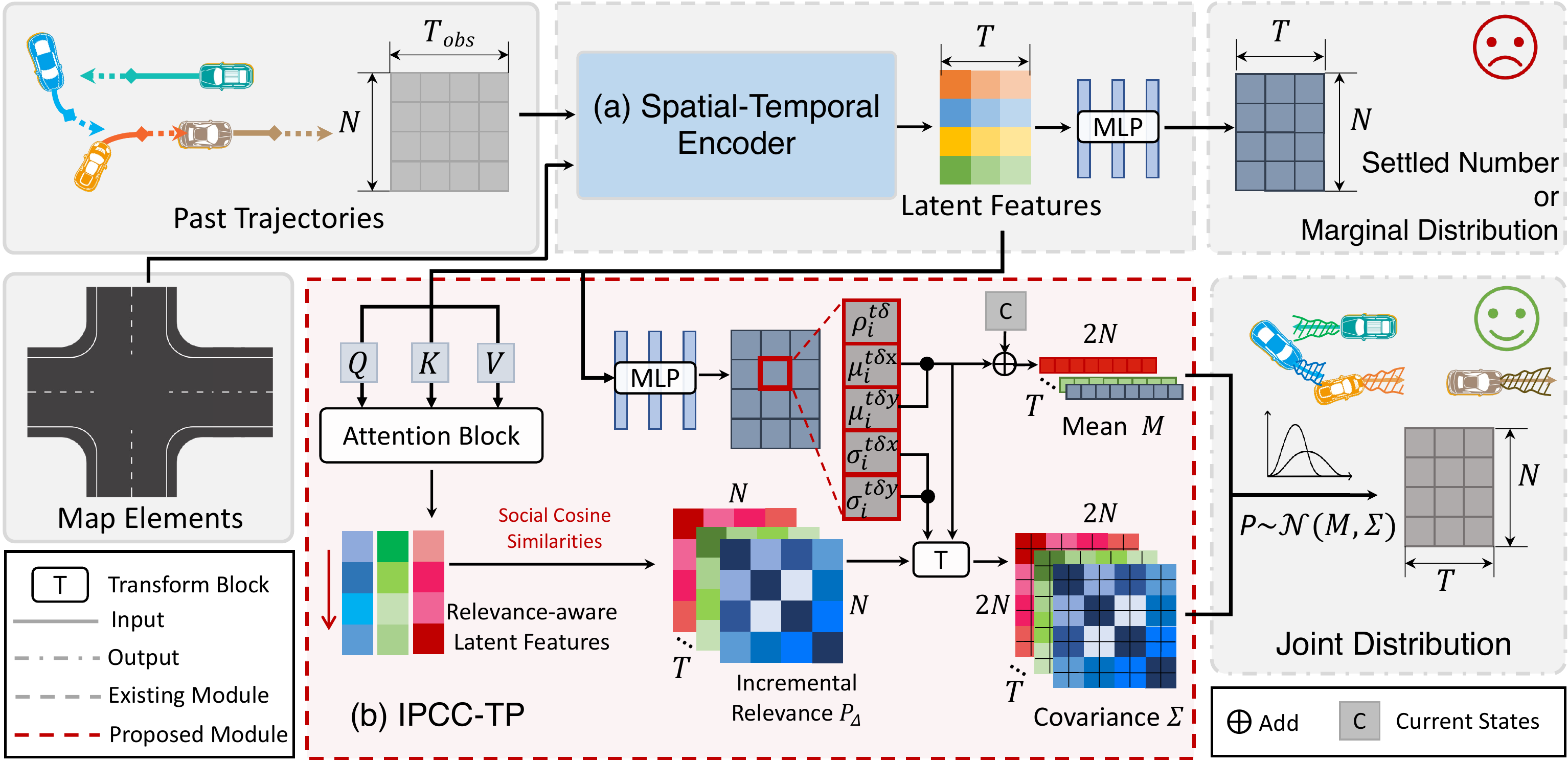}
    \caption{The overall framework of \textbf{IPCC-TP}. The inputs of existing MTP models include past trajectories and map elements. Latent features extracted by (a) Spatial-Temporal encoder are passed to (b) IPCC-TP to predict the future interactions represented by joint Gaussian distributions among the agents. IPCC-TP consists of an \textit{Attention Block} and a \textit{Transform Block}. Attention Block maps the original latent features to relevance-aware latent features, which are later used for calculating the incremental relevance $P_{\Delta}$ via cosine similarity measurement. The mean positions are obtained by adding the current positions of agents and the mean values in marginal distributions. Using incremental relevance $P_{\Delta}$ and $\mu^{tx}_i, \mu^{ty}_i, \sigma^{tx}_i, \sigma^{ty}_i$ in marginal distributions, Transform Block estimates the yaw angle and reconstructs the covariance matrices in the x-y coordinate system. 
   Defined by mean position $M$ and covariance matrices $\Sigma$, the joint Gaussian distributions $P\sim\mathcal{N}(M,\Sigma)$ for modeling the future interactions among the agents for $T$ steps are predicted by IPCC-TP.}
    \label{fig:pipeline}
\end{figure*}

\smallskip
\noindent
\textbf{IPCC Projection.}
Suppose that for agent $i$, the projections of mean displacement $\mu^{t \delta}_i$ on x- and y-axis, namely $\mu^{t \delta_x}_i$ and $\mu^{t \delta_y}_i$, are already obtained, we can estimate the approximate yaw angle for agent $i$ at step $t$ via
\begin{equation}
    \theta^t_i = \arctan(\mu^{t \delta_y}_i, \: \mu^{t \delta_x}_i)
    \label{eq:approx_yaw}.
\end{equation}
Similarly, $\Theta_t=\left[\theta_1^t, \ldots, \theta_N^t\right]$ at time $t$ is available.
The reconstructed future positions $\hat{\mathcal{F}_{t}^\ast}$ in the global coordinate can be expressed as a sum of the current states $M_{t_0}$ and increments.
$\hat{\mathcal{F}_{t}^\ast} \sim \mathcal{N}\left(M_{t}^\ast, \Sigma_{t}^\ast\right)$ is defined by
\begin{equation}
\begin{gathered}
\label{M*}
M_{t}^\ast = C \cdot M_{\Delta_r}^t+M_{t_0},\\[0.5em]
M_{\Delta_r}^t = \left[\mu^{t \delta}_1, \mu^{t \delta}_1, \ldots, \mu^{t \delta}_N, \mu^{t \delta}_N\right], \\[0.5em]
C = \operatorname{diag}\left(\left[\cos \theta_1^t, \sin \theta_1^t, \ldots, \cos \theta_N^t, \sin \theta_N^t \right]\right), 
\end{gathered}
\end{equation}
and
\begin{equation}
\setlength{\arraycolsep}{2pt}
\begin{gathered}
\Sigma_{t}^\ast = C^\mathsf{T} \Sigma_{\Delta_r}^t C=
\left[
\begin{array}{ccc}
{\sigma_{11}^{t \delta}} & \cdots & {\sigma_{1 N}^{t \delta}} \\
\vdots & \ddots & \vdots \\
{\sigma_{N 1}^{t \delta}} & \cdots & {\sigma_{N N}^{t \delta}}
\end{array}
\right],\\[1em]
\Sigma_{\Delta_r}^t = 
\left[
\begin{array}{ccccc}
\left(\sigma^{t \delta}_1\right)^2 & \left(\sigma^{t \delta}_1\right)^2 & \ldots & \sigma^{t \delta}_1 \sigma^{t \delta}_N & \sigma^{t \delta}_1 \sigma^{t \delta}_N\\
\left(\sigma^{t \delta}_1\right)^2 & \left(\sigma^{t \delta}_1\right)^2 & \ldots & \sigma^{t \delta}_1 \sigma^{t \delta}_N & \sigma^{t \delta}_1 \sigma^{t \delta}_N\\
\vdots & \vdots & \ddots & \vdots & \vdots\\
\sigma^{t \delta}_N \sigma_1^{t \delta} & \sigma^{t \delta}_N \sigma_1^{t \delta} & \cdots & \left(\sigma_N^{t \delta}\right)^2 & \left(\sigma_N^{t \delta}\right)^2\\
\sigma^{t \delta}_N \sigma_1^{t \delta} & \sigma^{t \delta}_N \sigma_1^{t \delta} & \cdots & \left(\sigma_N^{t \delta}\right)^2 & \left(\sigma_N^{t \delta}\right)^2
\end{array}\right],\\[1em]
{\sigma_{ij}^{t \delta}}=\rho^{t \delta}_{ij} \sigma^{t \delta}_i \sigma^{t \delta}_j \left[\begin{array}{cc}
 \cos{\theta_i^t} \cos{\theta_j^t} & \cos{\theta_i^t} \sin{\theta_j^t} \\[1em]
\sin{\theta_i^t} \cos{\theta_j^t} & \sin{\theta_i^t} \sin{\theta_j^t}
\end{array}\right],
\end{gathered}
\label{eq:sigma_tdij}
\end{equation}
where ${M_{\Delta_r}^t}$ and $\Sigma_{\Delta_r}^t$ are the replicated augments of $M_{\Delta}^t$ and $\Sigma_{\Delta}^t$ respectively, where $M_{\Delta_r}^t \in \mathbb{R}^{2 N}$ and $\Sigma_{\Delta_r}^t, C \in \mathbb{R}^{2 N \times 2 N}$. 
Note that when the approximate yaw angles $\Theta_t=\left[\theta_1^t, \ldots, \theta_N^t\right]$ are equal to the actual yaw angles $\Phi_t = \left[\phi_1^t, \ldots, \phi_N^t\right]$, $\hat{\mathcal{F}_{t}^\ast}$ and $\hat{\mathcal{F}_{t}}$ are identical distributions, which leads to: 

\begin{equation}
M_{t}^\ast = M_{t}, \quad \Sigma_{t}^\ast = \Sigma_{t},
\label{eq:identical_dist1}
\end{equation}
and
\begin{equation}
    \sigma^t_{i j} = \sigma^{t \delta}_{i j}.\\
    \label{eq:identical_dist2}    
\end{equation}
According to Eq.~\eqref{eq:joint_cov},~\eqref{eq:sigma_tdij},~\eqref{eq:identical_dist1}, and~\eqref{eq:identical_dist2},
we obtain $\rho_{ij}^{txx}, \rho_{i j}^{txy}, \rho_{i j}^{tyx}, \rho_{ij}^{tyy}$ with the signum function sgn($\cdot$),
\begin{small} 
\begin{equation}
\setlength{\arraycolsep}{2pt}
\left[\begin{array}{cc}
\rho^{t x x}_{ij}  & \rho^{t x y}_{ij} \\[1em]
\rho^{t y x}_{ij} & \rho^{t y y}_{ij}
\end{array}\right] = \rho_{i j}^{t \delta} \cdot \text{sgn}(\left[\begin{array}{cc}
\cos{\theta_i^t}\cos{\theta_j^t}  & \cos{\theta_i^t}\sin{\theta_j^t} \\[1em]
\sin{\theta_i^t}\cos{\theta_j^t} & sin{\theta_i^t}\sin{\theta_j^t}
\end{array}\right])
\label{eq:reconstruct_rho}
\end{equation}
\end{small} 

So far, we have shown that when the approximate yaw angles are equal to the actual yaw angles, $\rho_{ij}^{txx}, \rho_{i j}^{txy}, \rho_{i j}^{tyx}, \rho_{ij}^{tyy}$ can be obtained via Eq.~\eqref{eq:reconstruct_rho}.
For short-term trajectory prediction (no longer than 6s), the vehicles are unlikely to have a sharp turn in such a short time, thus the angle $\theta_i^t$ based on the incremental movement is close to the actual yaw angle $\phi^t_i$, and $\hat{\mathcal{F}_{t}^\ast}$ is a suitable approximation to $\hat{\mathcal{F}_{t}}$.

\subsection{IPCC-TP: a Plugin Module}
Based on the IPCC method, we design and implement a plugin module named IPCC-TP, which can be easily inserted into existing models. 
The pipeline of IPCC-TP is shown in Figure.~\ref{fig:pipeline}.
Previous methods use spatial-temporal encoders to extract latent features $\mathcal{L} \in \mathbb{R}^{N \times T \times d}$, which are subsequently sent to MLP decoders to generate deterministic predictions~\cite{yuan2021agentformer} or marginal distributions~\cite{girgis2021latent} (marked with the sad face). 
In contrast, our IPCC-TP first sends the latent features $\mathcal{L}_t \in \mathbb{R}^{N \times d}$ at time step $t$ to the \textit{Attention Block}.
This block is designated for modeling the future interactions via which the latent features $\mathcal{L}_t$ are mapped to the relevance-aware latent features $\mathcal{L}_t^{rel} \in \mathbb{R}^{N \times d}$.
Considering PCC is in the range of [-1, 1], we compute pairwise motion relevance based on cosine similarity.
In this way, if agent $i$ and agent $j$ share a higher motion relevance, their relevance-aware feature, $\mathcal{L}_{i t}^{rel} , \mathcal{L}_{j t}^{rel} \in \mathbb{R}^d$, will have a higher cosine similarity.
For future time step $t>0$, IPCC-TP computes the social cosine similarities as $P^t_\Delta$ in Eq.~\eqref{eq:Ptd} for each pair of agents. 
Then with the approximated yaw angles $\Theta_t$, $P^t_\Delta$ is used to predict the $\rho$ parameters in $\Sigma^t$ according to Eq.~\eqref{eq:reconstruct_rho}. 
Once the $\rho$ parameters in $\Sigma^t$ are obtained, together with \{$\mu^{tx}_i, \mu^{ty}_i, \sigma^{tx}_i, \sigma^{ty}_i, \rho^{txy}_{i}$\}, 
which are already available in the marginal probability distributions, we can extend the marginal probability distributions to a joint Gaussian distribution $\hat{\mathcal{F}_t}$ defined in Eq.~\eqref{eq:joint_mean} and Eq.~\eqref{eq:joint_cov}.
As shown in Figure.~\ref{fig:increment}, the approximated yaw angles are computed via Eq.~\eqref{eq:approx_yaw}.
In order to stabilize the training, we use Tikhonov Regularization~\cite{hilt1977ridge} in IPCC-TP by adding a small value $\Delta_{reg}$ to the main diagonal of covariance matrices $\Sigma_t$ to obtain $\Sigma_t^{reg}$, 
\begin{equation}
\Sigma_t^{reg} = \Sigma_t + \Delta_{reg} \cdot I,
\label{tik_reg_delta}
\end{equation}
where $I\in \mathbb{R}^{2 N \times 2 N}$, an identity matrix.

Thereby we adopted the following loss functions for training, following~\cite{girgis2021latent, yuan2021agentformer}.
In the training of the IPCC-TP embedded models, we replace the negative log-likelihood (NLL) loss of \cite{girgis2021latent, yuan2021agentformer} with the scene-level NLL loss. 
The scene-level NLL loss is calculated as
\begin{equation}
\begin{gathered}
\mathcal{L}_{NLL-scene} = \sum_{t=1}^T 0.5 [ln(|\Sigma_t|) + \\ 
(\mathcal{F}_t - M_t)^T \Sigma_t^{-1} (\mathcal{F}_t - M_t) + 2N ln(2\pi)].
\end{gathered}
\label{eq:scene_nll}
\end{equation}
Thus the loss function for training the enhanced AgentFormer is
\begin{equation}
\mathcal{L}^{Ag} = \mathcal{L}_{kld} + \mathcal{L}_{sample} + \mathcal{L}_{NLL-scene},
\end{equation}
where $\mathcal{L}_{kld}$ term is the Kullback-Leibler divergence for the CVAE~\cite{bowman2016generating} in AgentFormer, $\mathcal{L}_{sample}$ term is the mean square error between the ground truth and the sample from the latent $z$ distribution.
And the loss function for training the enhanced AutoBots is
\begin{equation}
\mathcal{L}^{Ab} = \mathcal{L}_{ME} + \mathcal{L}^{*}_{kld} + \mathcal{L}_{NLL-scene},
\end{equation}
where $\mathcal{L}_{ME}$ is a mode entropy regularization term to penalize large entropy of the predicted distribution, $\mathcal{L}^{*}_{kld}$ is the Kullback-Leibler divergence between the approximating posterior and the actual posterior. 

\subsection{Methodology Comparison to CUM}
\label{method_compare_cu}
CUM~\cite{tang2021collaborative} is another work that tries to predict the joint probabilistic distribution for the MTP task.
Compared with our IPCC-TP, the main difference lies in the interpretation and implementation of the covariance matrix.
In order to avoid the inverse calculation for covariance matrices $\Sigma_t$, CUM uses MLP decoders to predict all the parameters in $\Sigma_t^{-1}$. In contrast, our module instead predicts $\rho$ parameters in $P^t_\Delta$, which can intuitively encapsulate the physical meaning of motion relevance in future interactions, and then predicts $\Sigma_t$ based on Eq.~\eqref{eq:reconstruct_rho}. 
Experiments demonstrate the superiority of IPCC-TP.

\section{Experiments}
\label{experiments}

\subsection{Expermental Settings}

\smallskip
\noindent
\textbf{Datasets.}
We apply IPCC-TP to AgentFormer~\cite{yuan2021agentformer} and AutoBots~\cite{girgis2021latent} and evaluate the performance on two public datasets, \textit{nuScenes}~\cite{caesar2020nuscenes} and \textit{Argoverse 2}~\cite{Argoverse2}. nuScenes collects 1K scenes with abundant annotations for the research of autonomous driving.
For the prediction task, observable past trajectories are 2s long, and predicted future trajectories are 6s long.
The sample rate is 2 Hz. 
Argoverse 2 contains 250K scenes in total. 
In each scene, observable past trajectories are 5s long, and predicted future trajectories are 6s long. 
The sample rate is 10 Hz. 
Considering the vast number of scenes in Argoverse 2, we randomly pick 8K scenes for training and 1K for testing. 


\smallskip
\noindent
\textbf{Implementation Details.}
For both AgentFormer and AutoBots, we use the official implementations as the baselines in our experiments. 
For IPCC-TP, we implement the attention block inside it with a self-attention layer followed by a 2-layer MLP.
We set the Tikhonov regularization parameter $\Delta_{reg}$ as 1e-4 in the experiments. 
More details of the training are provided in the supplementary material.

\smallskip
\noindent
\textbf{Metrics.}
For evaluation, considering that the widely used metrics minADE and minFDE are designed for the ego-motion prediction in STP task, we choose to use \textit{Minimum Joint Average Displacement Error} (\textbf{minJointADE}) and \textit{Minimum Joint Final Displacement Error} (\textbf{minJointFDE}) in the multi-agent prediction task of INTERPRET Challenge~\cite{interactionchallenge,interactiondataset}. 
These two metrics are proposed specifically for the MTP task; their calculation formula is listed in the 
supplementary material. According to the rule of challenges on the nuScenes and Argoverse 2 benchmarks, we set the number of modes $M$ as 5 in the evaluation on nuScenes and 6 in the evaluation on Argoverse 2. 

\subsection{Training Details}

\smallskip
\noindent
\textbf{AgentFormer.} In the evaluation on nuScenes, we use the official pre-trained model as the baseline and the default hyperparameters for training the enhanced model. In the evaluation on Argoverse 2 dataset, since there are much more future steps and agents, we set the dimension as 24 and the number of heads as 4 in the attention layers to reduce VRAM cost. In both evaluations, we use Adam optimizer with a learning rate of 1.0e-4.

\smallskip
\noindent
\textbf{AutoBots.} In the evaluation on nuScenes, we use the default hyperparameters for AutoBots baseline and the enhanced AutoBots. In the evaluation on Argoverse 2 dataset, we set the dimension as 32 in the attention layers also for reducing the VRAM cost. In both evaluations, we use Adam optimizer with a learning rate of 7.5e-4. 

\subsection{Experimential Results}
We report and analyze the qualitative and quantitative results against AgentFormer and AutoBots on nuScenes and Argoverse 2.
After that, we compare the performance of CUM and IPCC-TP, and illustrate why our module can surpass CUM on the baseline. More results can be found in the supplementary material.

\begin{figure}[t]
    \centering
    \includegraphics[width=1.00\linewidth]{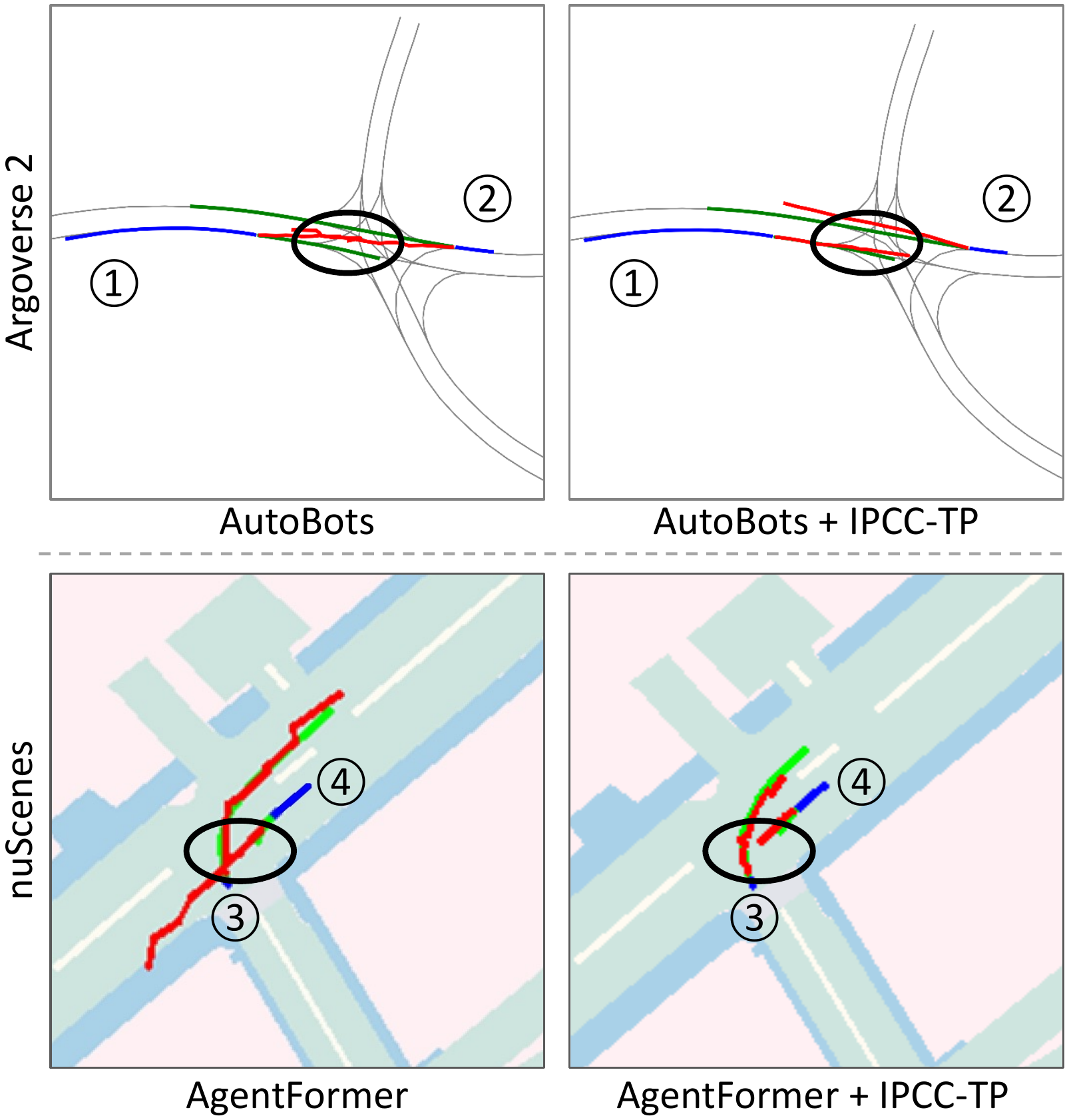}
    \caption{Comparisons between baselines and baselines + IPCC-TP on Argoverse 2 and nuScenes (blue line: past trajectories, green line: ground truth, red line: predictions). \textbf{Top}: Vehicle \ding{172} and \ding{173} drive towards different directions, and they should keep their lanes without interfering with each other. \textbf{Below}: Vehicle \ding{174} and \ding{175} drive aggressively, and \ding{175} is about to stop. Black circles show that the results of baselines are abnormal, while they become reasonable after plus IPCC-TP.}
    \vspace{-5mm}
    \label{fig:nuscenes_argo2}
\end{figure}

\smallskip
\noindent
\textbf{Evaluation on nuScenes.} We enhance AgentFormer and AutoBots with IPCC-TP for the evaluation on nuScenes. We use the official pre-trained AgentFormer as a baseline, which is reported to perform minADE(5) of $1.86m$ and minFDE(5) of $3.89m$ on nuScenes.
The experiment results are listed in Table.~\ref{tab:nuscenes} from which we can observe that IPCC-TP improves the original AgentFormer and AutoBots by $0.33m$ \bm{$(4.22\%)$} and $0.23m$ \bm{$(3.54\%)$} regarding minJointFDE(5) respectively. 
The bottom part of Figure.~\ref{fig:nuscenes_argo2} shows a case of comparisons between the original AgentFormer and the AgentFormer embedded with IPCC-TP, which illustrates that AgentFormer boosted by IPCC-TP can predict more compliant and logical future interactions with fewer collisions than the original one can do.
\begin{table}[t]
  \centering
  \begin{tabular}{@{}lcc@{}}
    \toprule
    Model & minJ-ADE(5) & minJ-FDE(5) \\
    \midrule
    AgentFormer~\cite{yuan2021agentformer} & 3.47 & 7.82 \\
    AgentFormer$+$IPCC-TP & \textbf{3.29} & \textbf{7.49} \\
    \midrule
    AutoBots~\cite{girgis2021latent} & 3.43 & 6.50 \\
    AutoBots$+$IPCC-TP & \textbf{3.22} & \textbf{6.27} \\
    \bottomrule
  \end{tabular}
  \caption{Evaluation on nuScenes.}
  \label{tab:nuscenes}
\end{table}
\begin{table}[t]
  \centering
  \begin{tabular}{@{}lcc@{}}
    \toprule
    Model & minJ-ADE(6) & minJ-FDE(6) \\
    \midrule
    AgentFormer~\cite{yuan2021agentformer} & 2.74 & 5.88 \\
    AgentFormer$+$IPCC-TP & \textbf{2.10} & \textbf{4.64} \\
    \midrule
    AutoBots~\cite{girgis2021latent} & 2.72 & 5.76 \\
    AutoBots$+$IPCC-TP & \textbf{2.20} & \textbf{4.79} \\
    \bottomrule
  \end{tabular}
  \caption{Evaluation on Argoverse 2.}
  \label{tab:argo2}
\end{table}

\begin{figure}[t]
    \centering
    \includegraphics[width=1.00\linewidth]{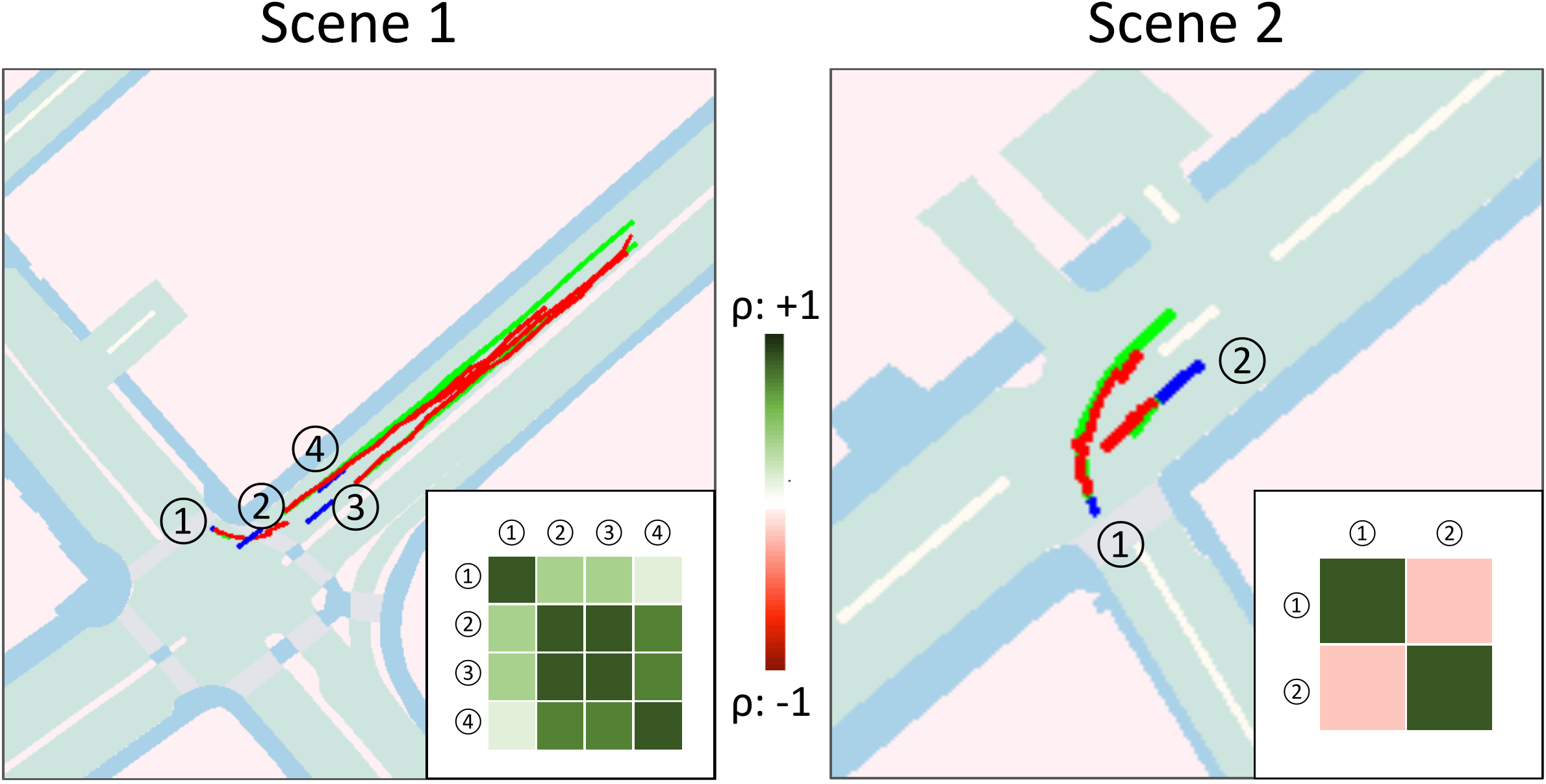}
    \caption{Visualization of correlation matrices $P^t_{\Delta}$ in 2 scenes of nuScenes dataset (blue line: past trajectories, light-green line: ground truth, red line: predictions, green area: driveable area). \textbf{Scene 1}: Vehicle \ding{173}$-$\ding{175} are about to leave the intersection while \ding{172} is just entering it. \ding{173}$-$\ding{175} are driving towards the same direction. \textbf{Scene 2}: Vehicle \ding{172} and \ding{173} meet at a T-junction. \ding{173} should yields to \ding{172} because the latter enters the junction first.}
    \vspace{-5mm}
    \label{fig:corr_viz}
\end{figure}

In Sec.~\ref{IPCC}, we have illustrated how $\rho$ in $P^t_{\Delta}$~\eqref{eq:Ptd} represents pairwise motion relevance. Here we show visualizations of $P^t_{\Delta}$ in different interaction patterns in Figure.~\ref{fig:corr_viz}.
In Scene 1, Vehicle \ding{172} is entering the intersection while Vehicle \ding{173}$-$\ding{175} have almost left the intersection, and they are driving with a similar speed towards the same direction.
IPCC-TP predicts $P^t_{\Delta}$ consisting of all positive values, and it also predicts a larger correlation between two vehicles that are closer to each other, e.g., \ding{172} has a larger correlation with \ding{173} than the correlation with \ding{175}.
In Scene 2, IPCC-TP predicts a negative correlation between two vehicles, which describes the interaction mode that \ding{173} is yielding to \ding{172}.

\smallskip
\noindent
\textbf{Evaluation on Argoverse 2.} We repeat the experiment on Argoverse 2 dataset with AgentFormer and AutoBots. 
The experiment results are listed in Table.~\ref{tab:argo2}. 
In this evaluation, IPCC-TP improves the original AgentFormer and AutoBots by $1.24m$ \bm{$(21.09\%)$} and $0.97m$ \bm{$(16.84\%)$} regarding minJointFDE(6) respectively.
One case for comparison between the original AutoBots and the AutoBots embedded with IPCC-TP is illustrated at the top part of Figure.~\ref{fig:nuscenes_argo2}. 
In this instance, AutoBots plus IPCC-TP predicts reasonable and collision-free interactions. 
In contrast, the predicted trajectories from the original AutoBots deviate significantly from the center lines, resulting in two vehicles crashing into each other. The prediction of collisions also reveals that the initial baseline is inefficient in modeling future interactions.

\smallskip
\noindent
\textbf{Quantitative Comparison to CUM.} We have compared CUM with our IPCC-TP methodologically in Sec.~\ref{method_compare_cu}. In this subsection, we conduct a quantitative comparison between these two plugin modules on both Argoverse 2 and nuScenes using AutoBots and AgentFormer as baselines respectively. 
The experiment results are listed in Table.~\ref{tab:sjtu_nips} and Table.~\ref{tab:sjtu_nips2}. 
In the evaluation, IPCC-TP is superior to CUM by \bm{$1.24\%$} on AutoBots and by \bm{$3.35\%$} on AgentFormer regarding minJointFDE.  
The reason is that, compared with CUM, IPCC-TP not only models the future interaction more intuitively but also boosts the performance of MTP models more efficiently with less memory cost. 

\begin{table}[t]
  \centering
  \begin{tabular}{@{}lcc@{}}
    \toprule
    Model & minJ-ADE(6) & minJ-FDE(6) \\
    \midrule
    AutoBots$+$CUM~\cite{tang2021collaborative} & 2.25 & 4.85 \\
    AutoBots$+$IPCC-TP & \textbf{2.20} & \textbf{4.79} \\
    \bottomrule
  \end{tabular}
  \caption{Comparison to CU-based module on Argoverse 2}
  \label{tab:sjtu_nips}
\end{table}
\begin{table}[t]
  \centering
  \begin{tabular}{@{}lcc@{}}
    \toprule
    Model & minJ-ADE(5) & minJ-FDE(5) \\
    \midrule
    AgentFormer$+$CUM~\cite{tang2021collaborative} & 3.43 & 7.75 \\
    AgentFormer$+$IPCC-TP & \textbf{3.29} & \textbf{7.49} \\
    \bottomrule
  \end{tabular}
  \caption{Comparison to CU-based module on nuScenes}
  \vspace{-3mm}
  \label{tab:sjtu_nips2}
\end{table}

\subsection{Ablation Study}
\label{tik_reg_mag}
Training the enhanced models involves the inverse computation of $\Sigma_t$ in \eqref{eq:scene_nll}.
In order to improve the robustness of the inverse computation, we use Tikhonov Regularization~\eqref{tik_reg_delta} on $\Sigma_t$ during training the models in the aforementioned experiments. 
In this subsection, we conduct an ablation experiment on the magnitude of Tikhonov Regularization parameter $\Delta_{reg}$~\eqref{tik_reg_delta}. 
Based on AutoBots, we repeat training the enhanced models on Argoverse 2 with $\Delta_{reg}$ in different magnitudes.
The experiment results are listed in Table.~\ref{tab:tikhonov}. Without any regulation, the training fails at the first iteration. When the regulation parameter is 1e-5, the training can run for the first epoch but still fails soon. In this experiment, 1e-4 is the most appropriate regulation we found for IPCC-TP. When we set the regulation to a relatively large value such as 1e-3, the performance of the enhanced model is even worse than the original model. We additionally conduct an ablation study on the Attention Block. More details can be found in the supplementary.

\begin{table}
  \centering
  \begin{tabular}{@{}lllcc@{}}
    \toprule
    $\Delta_{reg}$ & \quad & \quad & minJ-ADE(6) & minJ-FDE(6) \\
    \midrule
    1.0e-3 & \quad & \quad & 4.01 & 8.59 \\
    1.0e-4 & \quad & \quad & \textbf{2.20} & \textbf{4.79} \\
    1.0e-5 & \quad & \quad & 14.42 & 28.24 \\
    0 & \quad & \quad & - & - \\
    \bottomrule
  \end{tabular}
  \caption{Different parameter magnitudes in Tikhonov Regulation}
  \label{tab:tikhonov}
\end{table}

\section{Conclusion}
In this paper, we propose a new interaction modeling method named IPCC, which is based on one-dimensional motion increments relative to the current positions of the agents.
Compared with the traditional methods modeling the future interactions based on the x-y coordinate system, the IPCC method can intuitively reveal the pairwise motion relevance, and furthermore, with fewer parameters need to be predicted. 
Based on the IPCC method, we design and implement IPCC-TP. This module can be conveniently embedded into state-of-the-art MTP models to model future interactions more precisely by extending the original predictions in the format of marginal probability distributions to a joint prediction represented by a multivariate Gaussian distribution. 
Experiments demonstrate that IPCC-TP can boost the performance of MTP models on two real-world autonomous driving datasets.
{\small
\bibliographystyle{ieee_fullname}
\bibliography{egbib}
}

\newpage
\appendix
\section*{Supplementary Material of IPCC-TP}
\renewcommand{\thesubsection}{\Alph{subsection}}

\subsection{Metrics}
In our experiments, we leverage the \textit{Minimum Joint Average Displacement Error} (\textbf{MinJointADE}) and \textit{Minimum Joint Final Displacement Error} (\textbf{MinJointFDE}) as the evaluation metrics, which are specifically proposed for multi-agent trajectory prediction task in INTERPRET Challenge~\cite{interactionchallenge}. 
They are different from \textit{Minimum Average Displacement Error} (minADE) and \textit{Minimum Final Displacement Error} (minFDE), which are frequently used in the benchmarking of ego-motion prediction. 
Since we focus on predicting a scene-compliant future interaction among the agents, metrics that consider all agents in the scene, such as minJointADE and minJointFDE, are more suitable options. 

\smallskip
\textbf{MinJointADE} represents the minimum value of the Euclidean Distance averaged by time and all agents between the ground truth and the mode with the lowest value~\cite{interaction}. 
Note that in the problem statement in subsection 3.1, we only analyze the case when the number of modes $M=1$. 
In this case, the future states and the corresponding estimations at step $t$ are $\mathcal{F}_t = \{ \mathcal{F}^t_i | i=1,...,N\}$ and $\hat{\mathcal{F}_t} = \{ \hat{\mathcal{F}^t_i} | i=1,...,N\}$ respectively, 
where $\mathcal{F}^t_i = \{ \mathcal{F}^{t x}_i, \mathcal{F}^{t y}_i \}$ and $\hat{\mathcal{F}^t_i} = \{ \hat{\mathcal{F}^{t x}_i}, \hat{\mathcal{F}^{t y}_i} \}$ are the position ground truth and estimation for agent $i$ at this step. 
Since a specific current situation could develop into multiple possible future interactions, most MTP model predicts multiple interaction modes, where $M > 1$, thus
$\hat{\mathcal{F}^t_{i}} = \{ \mathcal{F}^t_{i m} | m=1, ..., M \}$
and
$\hat{\mathcal{F}^t_{i m}} = \{ \hat{\mathcal{F}^{t x}_{i m}}, \hat{\mathcal{F}^{t y}_{i m}} \}$.
The minJointADE is calculated as 

\vspace{-3mm}
\begin{footnotesize} 
\begin{equation}
    \text{minJ-ADE}=\min_{1 \leq m \leq M } \frac{1}{NT} \sum_{i,t} \sqrt{(\hat{\mathcal{F}^{t x}_{i m}} - \mathcal{F}^{t x}_i)^2 + (\hat{\mathcal{F}^{t y}_{i m}} - \mathcal{F}^{t y}_i)^2}.
\end{equation}
\end{footnotesize}

\textbf{MinJointFDE} represents the minimum value of the euclidean distance at the last predicted timestamps averaged by all agents between the ground truth and the mode with the lowest value~\cite{interaction}.
The minJointFDE is defined as 

\vspace{-3mm}
\begin{footnotesize} 
\begin{equation}
    \text{minJ-FDE}=\min_{1 \leq m \leq M } \frac{1}{N} \sum_{i} \sqrt{(\hat{\mathcal{F}^{T x}_{i m}} - \mathcal{F}^{T x}_i)^2 + (\hat{\mathcal{F}^{T y}_{i m}} - \mathcal{F}^{T y}_i)^2}.
\end{equation}
\end{footnotesize} 


\begin{figure}[t]
    \centering
    \includegraphics[width=1.00\linewidth]{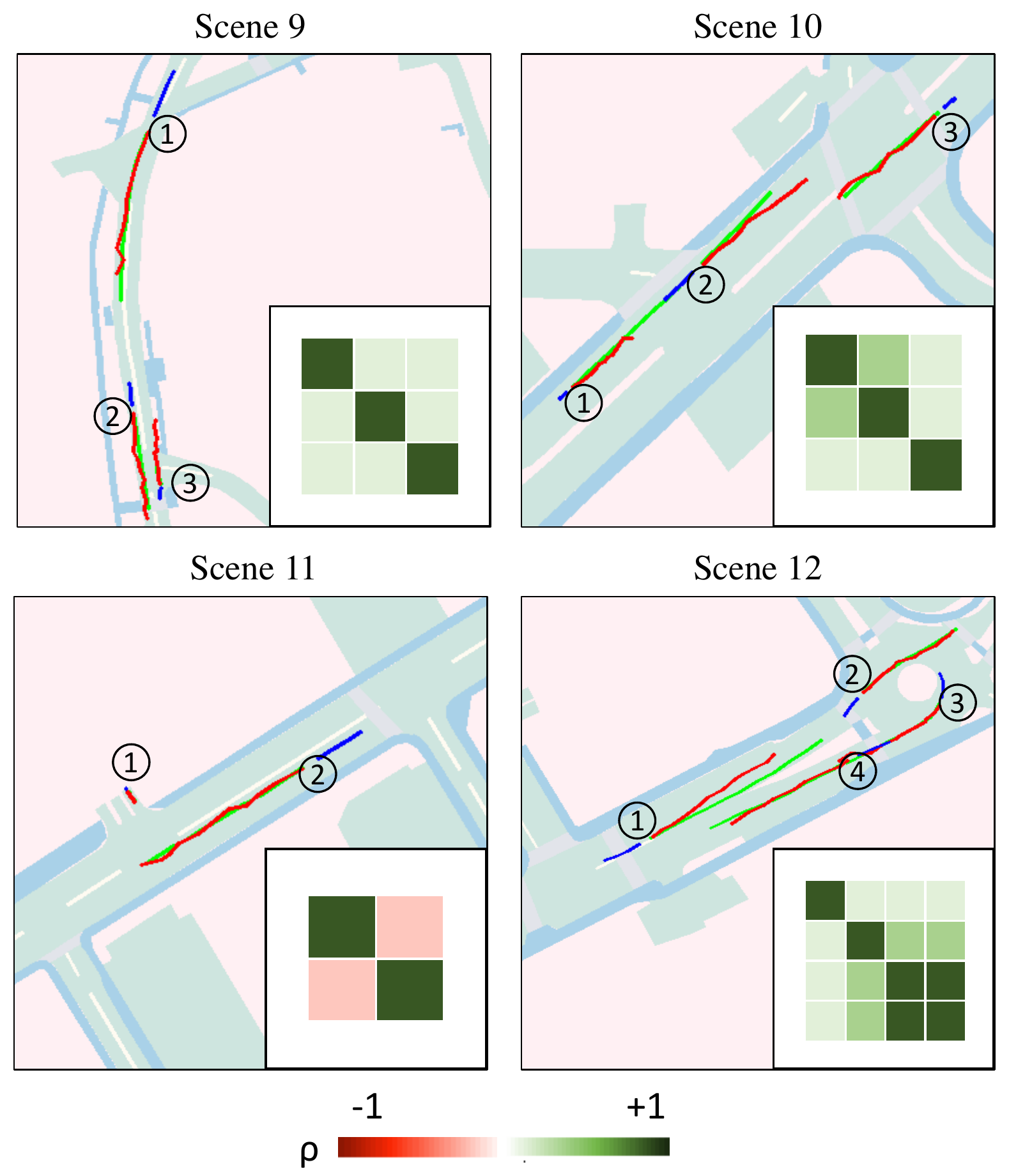}
    \caption{Visualization of IPCC matrices in nuScenes (blue line: past trajectories, green line: ground truth, red line: predictions). 
    \textbf{Scene 9}: Vehicle \ding{172} is far away from vehicle \ding{173} and \ding{174}, and the latter two vehicles are driving in opposite directions. \textbf{Scene 10}: Vehicle \ding{172} is following vehicle \ding{173}, and vehicle \ding{174} is driving on the opposite lane. 
    \textbf{Scene 11}: Vehicle \ding{172} is yielding to vehicle \ding{173}. 
    \textbf{Scene 12}: Vehicle \ding{172}, \ding{173}, \ding{174}, \ding{175} are 
   passing or approaching a roundabout, and vehicle \ding{174} is closely following vehicle \ding{175}.}
    \label{fig:corr_viz_2}
\end{figure}
\subsection{Experiment Details}
\subsubsection{Data Preprocessing}

\smallskip
\noindent
\textbf{NuScenes. }
In the nuScenes~\cite{caesar2020nuscenes} dataset, the observable histories and predicted futures respectively last for 2s and 6s, with a sample rate of 2Hz. Thus the histories contain 4 timesteps, and the futures contain 12 timesteps. In the data preprocessing for Agentformer~\cite{yuan2021agentformer} and AutoBots~\cite{girgis2021latent}, we remove agents with recorded past trajectories less than 2 steps or with incomplete future trajectories.

\smallskip
\noindent
\textbf{Argoverse 2. }
In the Argoverse 2~\cite{Argoverse2} dataset, the observable histories, and predicted futures are 5s and 6s with a sample rate of 10Hz. Henceforth, the histories contain 50 timesteps, and the futures contain 60 timesteps. Compared with nuScenes, Argoverse 2 has significantly more agents in most scenes, and most agents have complete future trajectories. Thus, we only select agents with complete trajectories in the data preprocessing for Agentformer and AutoBots.

\subsubsection{Training Details}
For IPCC-TP and the backbones of Agentformer and AutoBots, we set a dropout rate of 0.1. 
We train the Agentformer for 50 epochs with an initial learning rate of 1e-4 in the evaluation on both nuScenes and Argoverse 2. We further decay the learning rate by 0.5 every 10 epochs. As for the training of AutoBots, we set the initial learning rate as 7.5e-4. In the evaluation on nuScenes, we train the model for 100 epochs. 
Thereby, in the first 50 epochs, we decay the learning rate by 0.5 every 10 epochs. In the evaluation on Argoverse 2, we train the model for 100 epochs, reducing the learning rate by 0.5x every 20 epochs. We train the models on a single NVIDIA Titan Xp. 
\subsection{More Experiment Results}
In this supplementary material, we first provide more results about the qualitative comparisons between the baseline models and the enhanced models on nuScenes and Argoverse 2. Then, to support our statement about yaw angle $\theta$ estimation mentioned in Sec.3.3. IPCC Projection, we also study the error caused by the approximation in a quantitative way. Next, we provide the ablation study results on the Attention Block introduced in the main paper. Finally, we provide the result of AutoBots enhanced by our module compared with the original version on the INTERACTION dataset~\cite{interactiondataset}.
\subsubsection{Qualitative Results on nuScenes and Argo 2}
\begin{figure}[t]
    \centering
    \includegraphics[width=1.00\linewidth]{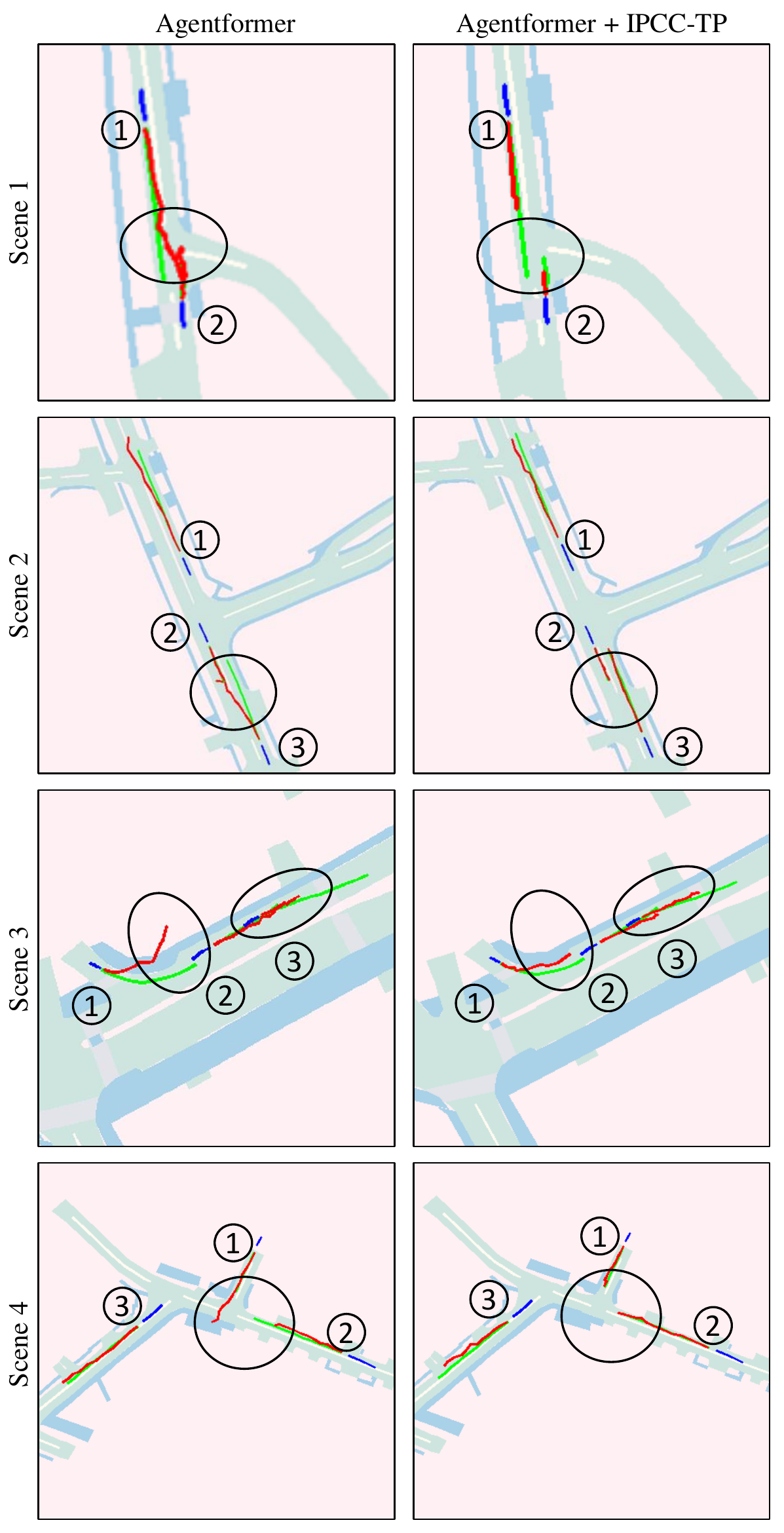}
    \caption{Comparisons between Agentformer and Agentformer + IPCC-TP on nuScenes (blue line: past trajectories, green line: ground truth, red line: predictions). \textbf{Scene 1}: Vehicle \ding{172} and \ding{173} drive in different directions, and they should keep their lanes without interfering with each other. \textbf{Scene 2}: Vehicle \ding{172}, \ding{173} and \ding{174} should keep their lanes.
    \textbf{Scene 3}: Vehicle \ding{172} should turn left properly and then follow vehicle \ding{173} and \ding{174}.
    \textbf{Scene 4}: Vehicle \ding{172} should wait until vehicle \ding{173} passes the T-junction.
    Black circles show that the results of the Agentformer baseline are abnormal, while they become reasonable after the model is enhanced with IPCC-TP.}
    \label{fig:nuscenes}
\end{figure}

\begin{figure}[t]
    \centering
    \includegraphics[width=1.00\linewidth]{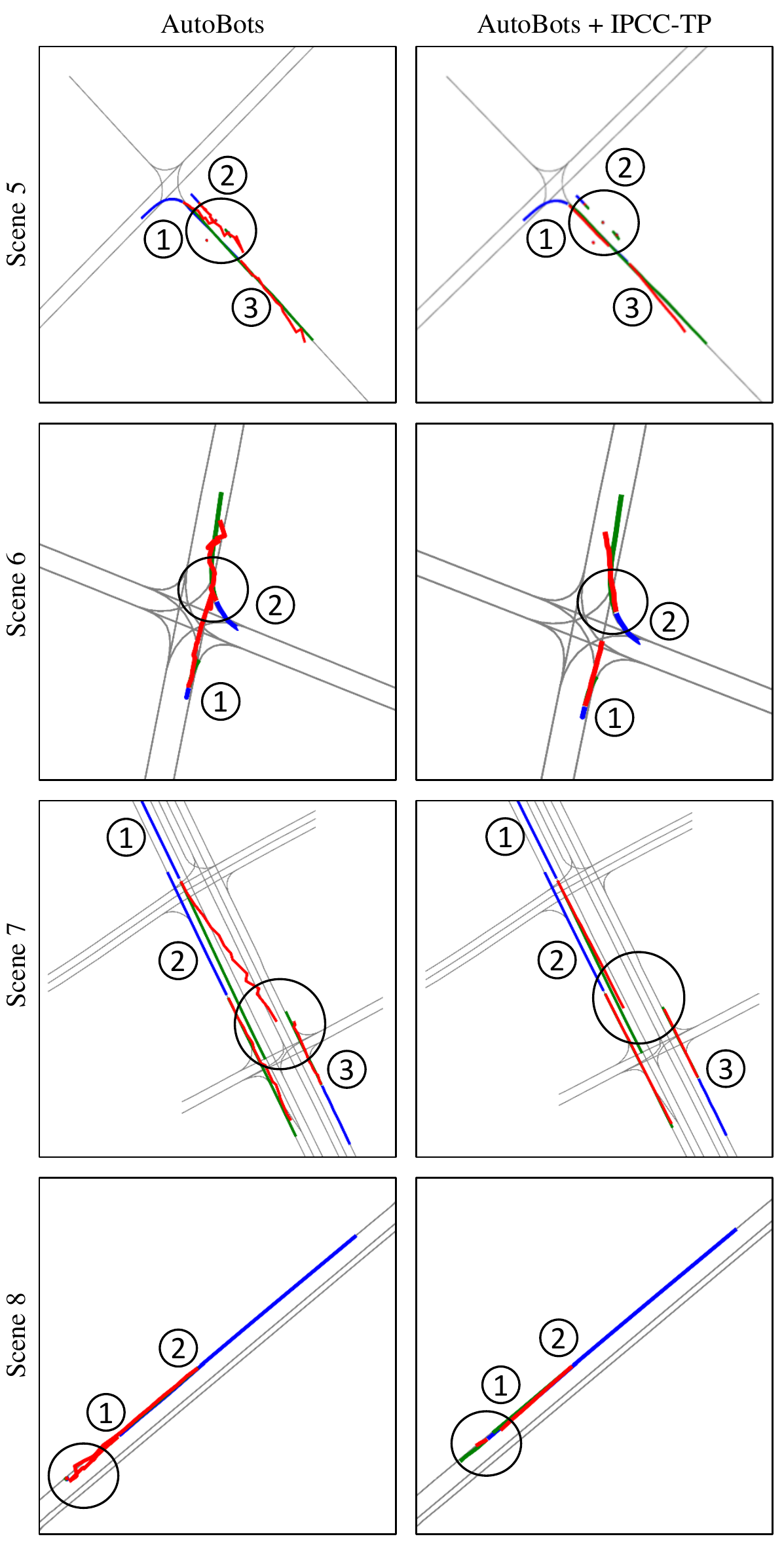}
    \caption{Comparisons between AutoBots and AutoBots + IPCC-TP on Argoverse 2 (blue line: past trajectories, green line: ground truth, red line: predictions, grey line: road center lines). \textbf{Scene 5}: Vehicle \ding{172} and \ding{174} should keep their lanes while vehicle \ding{173} is about to stop on the side of the road. 
    \textbf{Scene 6}: Vehicle \ding{173} is turning right and vehicle \ding{172} should yield to vehicle \ding{173}.
    \textbf{Scene 7}: Vehicle \ding{172}, \ding{173}, \ding{174} should keep their lanes without interfering with each other.
    \textbf{Scene 8}: Vehicle \ding{173} is approaching vehicle \ding{172} while the latter starts moving.
    Black circles show that the results of AutoBots baseline are abnormal, while they become reasonable after the model is enhanced with IPCC-TP.}
    \label{fig:argo2}
\end{figure}

We visualize IPCC matrices in several scenes in nuScenes. The result is shown in Figure.~\ref{fig:corr_viz_2}. In addition, we show more comparisons in Figure.~\ref{fig:nuscenes} and Figure.~\ref{fig:argo2}.  
\subsubsection{Yaw Angle Analysis}
\begin{figure}[t]
  \centering
  \includegraphics[width=\linewidth]{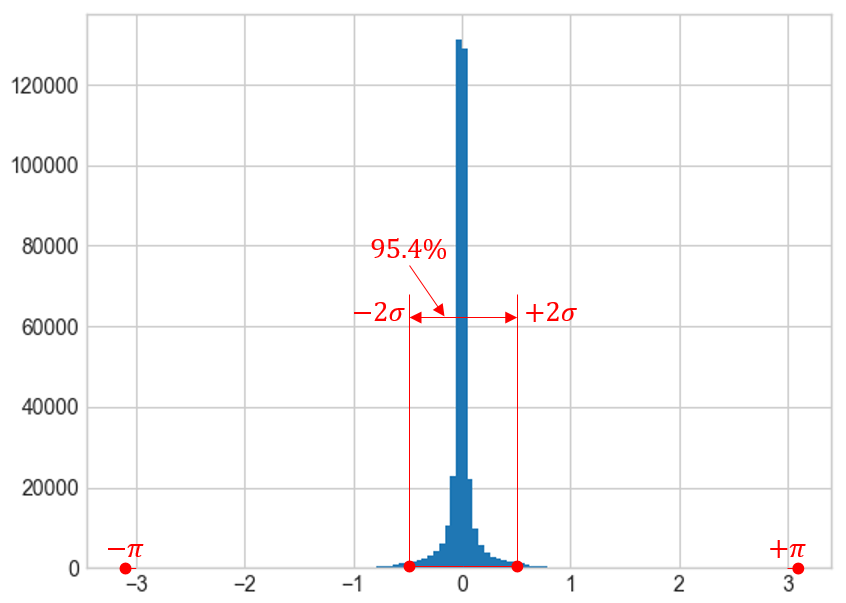}
  \caption{
  \textbf{Distribution of $\delta \theta$.} 
  According to 378k samples from the nuScenes dataset, the error $\delta \theta$ between the real yaw angle $\phi$ and the estimated yaw angle $\theta$ has a mean value of $-0.4^{\circ}$ and a standard deviation of $14.9^{\circ}$.}
  \label{fig:method:delta_error_dist}
\end{figure}

As described in Sec.3.3 in the main paper, for short-term trajectory prediction (no longer than 6s), vehicles are unlikely to have sharp turns in such a short time, thus the angle $\theta_i^t$ based on the incremental movement is close to the actual yaw angle $\phi^t_i$, and $\hat{\mathcal{F}_{t}^\ast}$ is a suitable approximation to $\hat{\mathcal{F}_{t}}$.
We counted the distribution of the error $\delta \theta$ between the real and estimated yaw angles from 378k samples.
It turns out that $\delta \theta \sim \mathcal{N}(-0.4, \ 14.9^2)$ (degree), which means that 95.4\% of the estimated angles have an error less than $30^{\circ}$ ($2\sigma$ rule), as depicted in Figure~\ref{fig:method:delta_error_dist}.
Thus, our approximation of $\theta$ is fairly reasonable.
\subsubsection{Ablation Study}
The proposed Attention Block is designed to assign the weight of others' influences to each agent. In this experiment, we substitute a Multi-Layer Perception (MLP) Block for it and summarize the result in Table~\ref{Ablation}. Results demonstrate the superiority of the Attention Block, which captures the cross-agent relevance.

\subsubsection{Quatitative results on INTERACTION} The INTERACTION dataset requires 3s predicted future trajectories, which is less challenging compared to nuScenes and Argoverse 2 (6s long). Thus, the demonstration of IPCC-TP's ability to model multi-agent relevance over long periods of time is limited when evaluated on INTERACTION. Regardless, we still provide our results in Table~\ref{tab:c} below for completeness. Although the baseline method AutoBots already demonstrates satisfactory results, IPCC-TP can still exceed its performance.
\begin{table}[t]
  \centering
  \scalebox{1}{
  \begin{tabular}{@{}lcc@{}}
    \toprule
   AutoBots+IPCC-TP & minJ-ADE(6) & minJ-FDE(6) \\
    \midrule
    MLP Block & 2.26 & 4.84 \\
    Attention Block & \textbf{2.20} & \textbf{4.79} \\
    \bottomrule
  \end{tabular}}
 \caption{Ablation study of Attention Block on Argoverse 2.}
  \label{Ablation}
\end{table}
\begin{table}[t]
  \centering
  \scalebox{1}{
  \begin{tabular}{@{}lcc@{}}
    \toprule
    INTERACTION & minJ-ADE(6) & minJ-FDE(6) \\
    \midrule
    AutoBots & 0.36 & 0.97 \\
    AutoBots+IPCC-TP & \textbf{0.35} & \textbf{0.93} \\
    \bottomrule
  \end{tabular}}
 \caption{Evaluation on INTERACTION.}
  \label{tab:c}
  \vspace{-2.5mm}
\end{table}

\end{document}